\documentclass[final]{cvpr}

\usepackage{times}
\usepackage{epsfig}
\usepackage{graphicx}
\usepackage{amsmath}
\usepackage{amssymb}
\usepackage{subfigure}
\usepackage{cite}

\usepackage[pagebackref=true,breaklinks=true,colorlinks,bookmarks=false]{hyperref}

\def\cvprPaperID{8229} 
\def\confYear{CVPR 2021}
\usepackage{microtype}
\usepackage{color}
\usepackage{colortbl}
\usepackage{multirow}
\usepackage{makecell}
\usepackage{hhline}
\usepackage{bbding}
\usepackage{float}
\usepackage{CJK}

\graphicspath{{figures/}}
\definecolor{lightgray}{gray}{.92}
\definecolor{tinygray}{gray}{.96}

\begin{document}
	
	\title{Learning Semantic Person Image Generation by Region-Adaptive Normalization}

	\author{Zhengyao Lv$^1$, Xiaoming Li$^1$, Xin Li$^2$, Fu Li$^2$, Tianwei Lin$^2$, Dongliang He$^2$, Wangmeng Zuo$^{1,3(}$\Envelope$^)$\\
		$^1$School of Computer Science and Technology, Harbin Institute of Technology, China\\
		$^2$Department of Computer Vision Technology (VIS), Baidu Inc.\\
		$^3$Pazhou Lab, Guangzhou, China\\
		{\tt\small cszy98@gmail.com \{csxmli, wmzuo\}@hit.edu.cn}
	}

	\maketitle
	\pagestyle{empty}  
	\thispagestyle{empty} 

	\begin{abstract}
		Human pose transfer has received great attention due to its wide applications, yet is still a challenging task that is not well solved. Recent works have achieved great success to transfer the person image from the source to the target pose. However, most of them cannot well capture the semantic appearance, resulting in inconsistent and less realistic textures on the reconstructed results. To address this issue, we propose a new two-stage framework to handle the pose and appearance translation.  
		In the first stage, we predict the target semantic parsing maps to eliminate the difficulties of pose transfer and further benefit the latter translation of per-region appearance style. In the second one, with the predicted target semantic maps, we suggest a new person image generation method by incorporating the region-adaptive normalization, in which it takes the per-region styles to guide the target appearance generation. Extensive experiments show that our proposed SPGNet can generate more semantic, consistent, and photo-realistic results and perform favorably against the state of the art methods in terms of quantitative and qualitative evaluation.  The source code and model are available at \url{https://github.com/cszy98/SPGNet.git}.
	\end{abstract}
	
	\vspace{-5pt}
	\section{Introduction}
	\vspace{4pt}
	Given an image of a specific person under a certain perspective or pose, the human pose transfer task aims to generate images of the {person with the same appearance and meanwhile under the given target} pose, which {more importantly,} are expected to be as photo-realistic as possible. The task has a wide range of applications, such as generating videos with pose sequences~\cite{balakrishnan2018synthesizing,2019Coordinate,2019Pose,liu2019liquid,Yang_2018_ECCV}, data augmentation for person re-identification task~\cite{2018Pose,zhu2019progressive} {and multi-view display for virtual try-on~\cite{2017The,Yu2019VTNFP,Yang2020Towards}.}
	
	{Basically, human pose transfer is a non-rigid deformation of the 3D human. Only using one 2D source image and target pose to generate another view of the human body is still a challenging task due to the following difficulties: (i) spatial rearrangement of the appearance features, (ii) the inference of the self-occlusion region, (iii) the photo-realistic results. All these make the task valuable and remain an active topic in the community of computer vision.}
	
	{On the one hand, along with the rapid development of deep learning, cGAN~\cite{mirza2014conditional} based human pose transfer methods~\cite{ma2017pose,Pumarola_2018_CVPR,tang2019cycleincycle} have achieved significant progress. However, this task is an unaligned image to image translation due to the inconsistent poses between the source and target {images}. Directly taking the concatenation of the source appearance and target pose into a general encoder-decoder framework cannot fully exploit the correspondence between the source and target {appearance}, thus resulting in less realistic results.}
	{On the other hand, to facilitate the spatial rearrangement of the source appearance, deformation~\cite{siarohin2018deformable,li2019dense,liu2019liquid,ren2020deep} and disentanglement~\cite{esser2018variational,men2020controllable} based methods have been suggested. Albeit the appearance alignment problem can be addressed, the generated distortion and blurry appearance caused by the warped and disentangled features are inevitable, giving rise to visually unpleasing results. The recent progressive generation~\cite{zhu2019progressive, tang2020xinggan} also achieved plausible performance, but the final results usually lose semantic details. {Though  improvements have been obtained, the semantic generation remains uninvestigated reasonably in the human pose transfer task.}}
	
	{In this paper, we present a two-stage {framework} to address the above issues. In the first stage, we predict the target semantic parsing maps. Compared with these directly generating the target image, the prediction of target parsing maps is much easier, because we do not need to consider the texture {translation} and this will make the network focus on only one single task, \ie, pose transfer without considering the effect of appearance that may bring the burden to the learning of the network. {With the given pose and source parsing map, we suggest a SPATN model to generate the target semantic maps in a progressive manner.} We observe that the target semantic maps can bring the following benefits: 
		(i) it can not only provide the pose information, but also the {specification} of each body region, making our model more robust when dealing with complex poses, (ii) with the per-region styles, it can help to transfer them to the target region separately, which is more effective in generating the semantic and photo-realistic results. {In the second stage, we propose a SPGNet by incorporating the semantic region adaptive normalization to generate the target image.} To be specific, it takes the per-region styles that are extracted from the source appearance and then broadcasted to the target regions to assist the semantic generation on each body part, resulting in the final photo-realistic results. This manner can well address the inconsistent poses and self-occlusion problems. We note that there are also some works that attempt to utilize the predicted target parsing maps to improve the generation of the final results~\cite{dong2018soft,song2019unsupervised,men2020controllable}. However, Dong \etal~\cite{dong2018soft} and Song \etal~\cite{song2019unsupervised} both directly take the predicted target parsing map and source image as input. This is more like a cGAN, which can bring limited improvements {to} the final semantic reconstruction. Men \etal~\cite{men2020controllable} also suggest the extraction of per-region style codes, but these codes are utilized in the reconstruction process by the AdaIN layer~\cite{huang2017arbitrary}, {in which the learned affine parameters from the source appearance are uniform across spatial coordinates}. We argue that compared with the global normalization, the region adaptive normalization is more flexible and suitable for this task {by specifying the style to each target region.}}
	
	{Experiments are conducted to evaluate the effectiveness of our proposed SPGNet on two challenging datasets, \ie, DeepFashion~\cite{liu2016deepfashion} and Market-1501~\cite{zheng2015scalable}. Results show that our SPGNet performs favorably against the state of the arts, and yields more consistent and photo-realistic results. The main {contributions} of this work include:}
	\begin{itemize}
		\item {We predict the target semantic maps as a supplement to the pose representation, and then further utilize it to the per-region style translation by region adaptive normalization, thereby eliminating the difficulties of the pose transfer and improving the reconstruction quality.}
		\item {In comparison with these state of the arts methods, experiments demonstrate the superiority of our SPGNet in generating favorable and photo-realistic person image.}
	\end{itemize}
	\vspace{6pt}
	\section{Related Work}
	\vspace{2pt}
	\subsection{Human Pose Transfer}
	
	{The task of human pose transfer has achieved great development during these years, especially with the unprecedented success of deep learning. Originally, these methods regard this task as a conditional image generation by taking the source appearance image as a condition to guide the generation of the target image, which mainly comes from the conditional generative adversarial networks (cGANs)~\cite{mirza2014conditional}. Ma \etal~\cite{ma2017pose} first present a two-stage model, which directly concatenated the target pose and the source image as input to generate the target image in a {coarse-to-fine} manner. Pumarola \etal~\cite{Pumarola_2018_CVPR} suggest {an unsupervised} manner by taking the concatenation of the generated results and the source pose as input to reconstruct the source image. Similarly, Tang \etal~\cite{tang2019cycleincycle} propose a {cycle-in-cycle} way to constrain the learning of pose and appearance translation. However, directly concatenating the target pose and source image usually {brings} limited improvements due to the inconsistent poses. To solve the above issues that hinder the appearance translation, Essner \etal~\cite{esser2018variational} adopt the combination of VAE~\cite{kingma2013auto} and U-Net~\cite{isola2017image} to disentangle the appearance and the pose of person images. Men \etal~\cite{men2020controllable} also adopt the disentanglement by extracting the region styles to perform on the whole target pose. Siarohin \etal~\cite{siarohin2018deformable} proposed a Deformable GAN, which decomposes the overall deformation by a set of local affine transformations to deal with the misalignments caused by different poses. Subsequently, in order to enhance the spatial rearrangement ability, there are also some works ~\cite{li2019dense,liu2019liquid,ren2020deep} that use flow-based deformation to align the source appearance. Both \cite{li2019dense} and \cite{liu2019liquid} adopt additional 3D human models to calculate the flow fields between the source and target image, while Ren \etal~\cite{ren2020deep} obtain the global flow fields in an unsupervised manner and further propose a local neural texture render. Recently, Zhu \etal~\cite{zhu2019progressive} propose to progressively transform the source image by a sequence of {pose-attentional transfer blocks (PATBs)}, which is flexible but useful information may be lost during multiple transfers. Based on these observations of this work, Tang \etal~\cite{tang2020xinggan} {further} propose a XingGAN model, which consists of shape-guided appearance-based generation branch, appearance-guided shape-based generation branch as well as co-attention fusion module to effectively transfer and update person shape and appearance features in a crossing and progressive manner. However, nearly all  these methods directly utilize the global appearance features on the target one, and seldom explore the benefits of per-region appearance translation that may bring to the final reconstruction.}

	{On the other hand, most of these aforementioned methods use human keypoints as pose representation due to its cheapness. However, it ignores the body skeleton that is useful to build the human body. There are also some other works that use DensePose\cite{neverova2018dense} or semantic parsing maps~\cite{dong2018soft,song2019unsupervised} as pose representation, which can provide more information about body depth or part segmentation. However, they directly take the semantic maps as a condition that is concatenated to the pose image along the channel dimension. Therefore, they can achieve only limited improvements on the target results especially the {semantic appearance of each region} due to the misalignment. Here, we also use {the target semantic parsing map} to assist the image generation but utilize it through the region adaptive normalization {on each body region} to separately guide the semantic appearance generation. }
	
	\subsection{{Semantic Image Generation}}
	
	Semantic image generation aims at synthesizing photo-realistic images {from the given semantic layout}. Several methods~\cite{isola2017image,wang2018high,park2019semantic,liu2019learning,zhu2020sean,tang2020local} have been suggested for solving this task. SPADE~\cite{park2019semantic} adopts semantic label maps to predict the spatially-varying affine transformation parameters {for incorporating the class prior to the target image}, which controls the image generation process more precisely and obtains visual fidelity results. 
	up{Similarly, CC-FPSE~\cite{liu2019learning} proposes to predict the spatially-varying conditional convolutional kernels based on the input semantic layout. 
		LGGAN~\cite{tang2020local} takes global and local contexts into account and gets promising performance. The image-level global generator learns a global appearance distribution and the class-specific local generator generates different object classes separately. 
		Furthermore, SEAN~\cite{zhu2020sean} improves SPADE by introducing per-region style encoding, which is better suited to encode, transfer, and synthesize style in terms of visual quality. Inspired by the benefit brought by SEAN~\cite{zhu2020sean}, in this work, we handle the person image generation by separate per-region style translation. To be specific, we adopt the {region adaptive normalization} {on each semantic region} to retain and transfer the style code between each corresponding semantic part of the source and target images.
	}
	
	\section{Method}
	\begin{figure*}[!t]
		\centering
		\includegraphics[width=.92\linewidth]{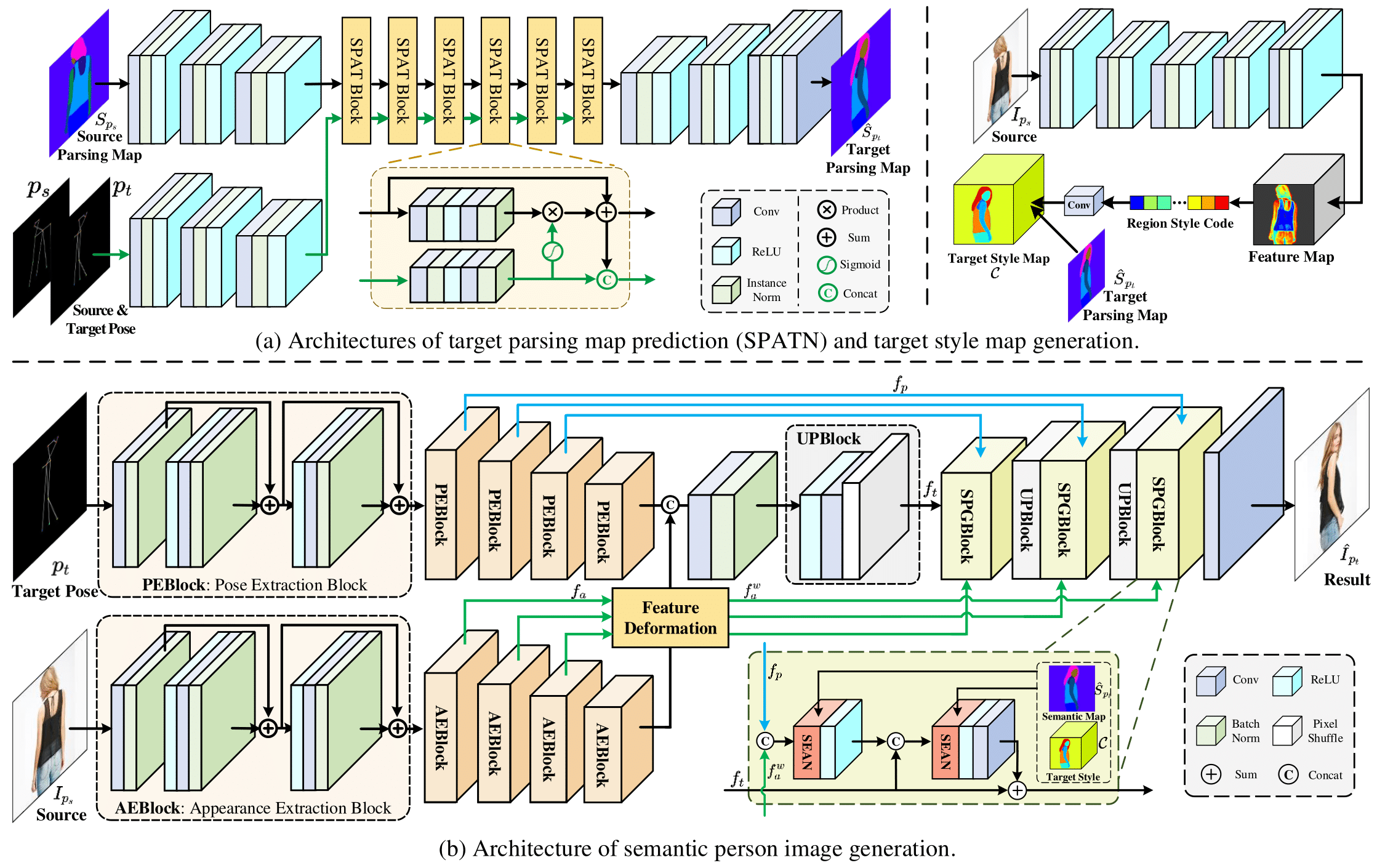}
		\caption{{Overview of our proposed method. It mainly contains two stages. In the first one, SPATN is adopted to generate the target parsing map $\hat{S}_{p_t}$. In the second one, feature deformation is utilized to warp the source appearance feature to the target pose, and SPGBlock is progressively introduced to incorporate the semantic region adaptive normalization on the generation of the target result $\hat{I}_{p_t}$.}}
		\label{fig:pipeline}
		\vspace{-12pt}
	\end{figure*}
	Given the source person image $I_{p_s}$ in pose $p_s$ and target pose $p_t$, our goal is to generate a photo-realistic image $\hat{I}_{p_t}$, which {has} the consistent appearance with $I_{p_s}$ while under the pose $p_t$. {Due to the large changes of views and the unseen regions caused by {self-occlusion}, directly predicting the desired result tends to be intractable. To address this issue, we propose a two-stage framework to eliminate these difficulties. In the first stage, we generate the semantic parsing maps $\hat{S}_{p_t}$ of the target pose from the source parsing maps $S_{p_s}$ , source pose $p_s$ and target pose $p_t$ in a progressive manner.
		In the second one, we use the predicted target semantic parsing maps $\hat{S}_{p_t}$ to guide the target image generation by the semantic region-adaptive normalization layers.
		The overall framework of the method is shown in Fig. \ref{fig:pipeline}.}
	%
	\subsection{{Target Semantic Parsing Map Generation}}
	{Instead of directly predicting the target person image {$\hat{I}_{p_t}$}, the prediction of {the} target parsing map tends to be much easier due to the lack of texture translation. However, it still {suffers from} the large pose changes between the source and the target parsing maps. {Inspired by the PATN~\cite{zhu2019progressive}, which proposed a progressive manner to handle the pose image transfer, here we adopt a semantic pose-attentional transfer network (SPATN) to model the semantic translation. The framework is shown in the left part of Fig.~\ref{fig:pipeline} (a).} It takes the source pose $p_s$, target pose $p_t$ and source semantic map $S_{p_s}$ as input to predict the target semantic map $\hat{S}_{p_t}$, which can be formulated as:
		\begin{equation}
		\setlength{\abovedisplayskip}{5pt}
		\setlength{\belowdisplayskip}{5pt}
		\hat{S}_{p_t} = \mathcal{F}_{SN}(p_s, p_t, S_{p_s}; \Theta_{SN}),
		\end{equation}
		where $\mathcal{F}_{SN}$ and $\Theta_{SN}$ denote the proposed SPATN model and its learnable parameters, respectively.}

	{We observe that using sparse keypoint to represent the pose can only provide limited body structure information and it ignores the correspondence of each part, which usually fails to deal with some complex poses (\eg crossed arms).
		In order to better model the pose structure and further benefit the generation of target semantic parsing,} we introduce the distance map as a complementary representation of pose structure. 
	{Originally, the pose $p$ is constructed by 18-channels heat maps that each one encodes one joint of a human body. With these 18 points, we generate 12 lines \{$L_m|_{m=1}^{12}$\} to represent the body skeleton.
		In this work, each skeleton generates one channel distance map. When this skeleton is invisible or occluded, the distance map is set to 0. 
		Thus we can generate {a 12-channels distance map} $\{M_m|_{m=1}^{12}\}$, which has the same width and height {as} the source image. 
		The values in $(x, y)$ of each $M_m$ is calculated by the smallest distance between the point $(x, y)$ and the skeleton $L_m$. The m-\textit{th} distance map can be obtained by:
	}
	\begin{equation}
	\setlength{\abovedisplayskip}{5pt}
	\setlength{\belowdisplayskip}{5pt}
	M'_m(x,y) =\underset{(x',y')\in L_m}{\min}\{\sqrt{(x-x')^2+(y-y')^2}\},
	\end{equation}
	{where $(x',y')$ denotes the point on the skeleton $L_m$.}
	{Here, we further normalize these values by introducing a negative parameter {$\kappa$}. The final distance map is then defined as:}
	\begin{equation}
	\setlength{\abovedisplayskip}{5pt}
	\setlength{\belowdisplayskip}{5pt}
	M_m(x,y) = exp (\kappa * M'_m(x,y)),
	\end{equation}
	{where $m \in \{1,2,3...12\}$ represents the m-\textit{th} skeleton.}
	{In this way, the closer the point to the skeleton, the larger the value is. Thus it can well model the body structure. In this work, $\kappa$ is set to $-0.1$. 
	}
	{Finally, the poses $p_s$ and $p_t$ {can be represented by} 30-channels features, respectively, which consist of 18-channels {joint} heat maps and the extra 12-channels distance maps.}
	up{The illustration and analyses of the generated distance map $M$ can be found in our suppl.} 
	
	\subsection{Target Image Generation}
	{Fig.~\ref{fig:pipeline} (b) illustrates the framework of our proposed semantic person generation network (SPGNet). It takes the target pose $p_t$, source image $I_{p_s}$, as well as target semantic parsing map $\hat{S}_{p_t}$ (predicted through SPATN in the first stage) as input to generate a photo-realistic target image $\hat{I}_{p_t}$, which has the same appearance with $I_{p_s}$ and the same pose with $p_t$. The whole model can be formulated as:
		\begin{equation}
		\setlength{\abovedisplayskip}{5pt}
		\setlength{\belowdisplayskip}{5pt}
		\hat{I}_{p_t} = \mathcal{F}_{SPG}(p_t, I_{p_s}, \hat{S}_{p_t}; \Theta_{SPG}),
		\end{equation}
		where $\Theta_{SPG}$ denotes the SPGNet model parameters.
	}
	
	{Following~\cite{siarohin2018deformable,zhao2018multi,li2019dense}, we adopt a dual-path UNet~\cite{ronneberger2015u} to encode the appearance and pose features separately. {The pose features are extracted by the suggested PEBlocks, while the appearance features are obtained through the AEBlocks (see Fig. \ref{fig:pipeline} (b)).} The PEBlock and AEBlock have the same architectures, except the channel numbers in the first block (30 for PEBlock and 3 for AEBlock). Each block is constructed by two residual blocks~\cite{he2016deep}. Inspired by~\cite{li2019dense}, here we adopt a feature deformation operation that is performed on the appearance features $f_a$ to solve the inconsistent poses. With the warped appearance feature $f_a^w$ and target pose feature $f_p$, we further propose a semantic person generation block (SPGBlock) by the region-adaptive normalization layers. The final result is generated by several stacks of UpBlock and SPGBlock in multi-scale feature spaces.}
	
	\noindent\textbf{Feature Deformation.} 
	{Several works have been proposed to address the appearance warping problem~\cite{balakrishnan2018synthesizing,siarohin2018deformable,li2019dense,ren2020deep}, which usually learn the global or local spatial transformation. Among these methods, Intr-Flow~\cite{li2019dense} proposed to learn the dense and intrinsic 3$D$ appearance flow by fitting a 3$D$ body model, which is more suitable to the pose transfer task. With the visibility map and the accurate flow prediction, this model can well handle the deformation of the appearance feature. Denote by the feature deformation module $\mathcal{F}_W$, the input appearance feature $f_a$. Following~\cite{li2019dense}, the warped appearance feature $f_a^w$ can be formulated as:}	
	\begin{equation}
	\setlength{\abovedisplayskip}{5pt}
	\setlength{\belowdisplayskip}{5pt}
	f_a^w = \mathcal{F}_W(f_a, \Phi_{2D}, V; \Theta_w),
	\end{equation}
	{where $\Phi_{2D}$ is obtained from the projection of the predicted $3D$ flow and $V$ is the visibility map. Both of them are generated by the pre-trained Intr-Flow model~\cite{li2019dense}. $\Theta_w$ is the learnable parameters of the deformation module $\mathcal{F}_W$. We adopt the same setting as~\cite{li2019dense} by incorporating a gating layer to manually exclusive {the visible and invisible} regions based on the visibility map $V$. {More details can be referred to our supplementary material}. In this work, we utilize the flow regression module of Intr-Flow~\cite{li2019dense}.
	}
	
	\noindent\textbf{Semantic Person Image Generation.} 
	{Basically, except for the pose transfer, keeping the appearance texture is also important for the person image generation task. However, recent works usually cannot well capture the appearance and then fail to transfer to their target semantic regions. We observe that even though the source and target appearance have somewhat differences due to the large pose changes, their corresponding semantic regions should share the same appearance. For instance, the color or texture of the target clothes which can be regarded as a kind of style should be consistent with the source one. Inspired by the SEAN~\cite{zhu2020sean} which proposed a semantic region adaptive normalization to individually control the style of each semantic part, we utilize it in our SPGNet to facilitate the appearance translation. To obtain the per-region styles, here we firstly adopt a general encoder-decoder network by taking the source image $I_{p_s}$ as input (as shown in the right part of Fig.~\ref{fig:pipeline} (a)). For each semantic region, we then generate the per-region style code through a region-wise average pooling layer on the feature map. Each style code is expected to contain the appearance features for each body region. 
		After obtaining the region style codes, we adopt several convolution layers to further encode each one by {excluding the effect of the other irrelevant regions}. With the predicted target semantic map $\hat{S}_{p_t}$ which is generated by SPATN in the first stage, we further broadcast the style code to their target region to generate the target style map $\mathcal{C}$(see Fig. \ref{fig:pipeline} (a)).
	}
	
\begin{figure}[t]
		\centering
		\includegraphics[width=.99\linewidth]{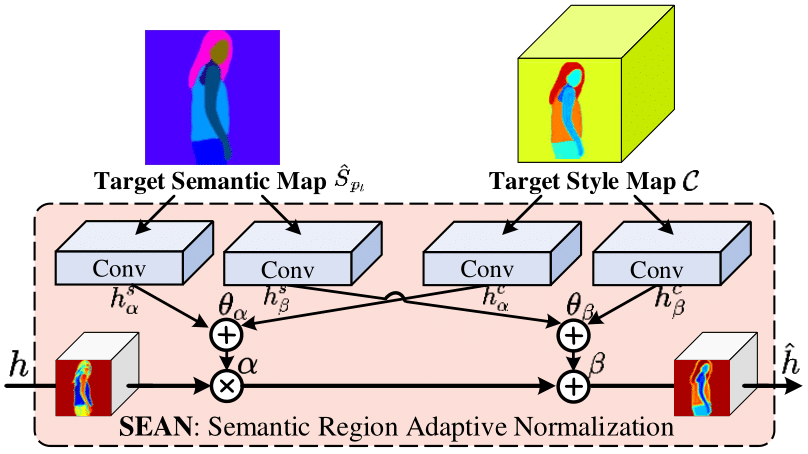}
		\caption{The details of the SEAN for semantic person generation.}
		\label{fig:sean}
		\vspace{-13pt}
	\end{figure}

	{With the pose feature $f_p$, the warped appearance feature $f_a^w$, the decoder feature $f_t$, the predicted target semantic map $\hat{S}_{p_t}$ and the target style map $\mathcal{C}$, we suggest a {semantic person generation block (SPGBlock)} by incorporating the semantic region normalization to generate the target person feature. 
		{As is shown in Fig.~\ref{fig:pipeline} (b), each block contains two SEAN operations.}
		The first one takes the $f_p$ and $f_a^w$ as input to generate the preliminary target feature. Then the second one takes the former output and the decoder feature $f_t$ to generate the residual features of the $f_t$. The output of the {SPGBlock} is expected to fuse the appearance to the pose features on each semantic region, respectively. The details of SEAN are shown in Fig.~\ref{fig:sean}. 
	}
	
	{Denote by the input feature as $h$, the output of SEAN in the position $(n,c,y,x)$ is given by :
		\begin{equation}
		\setlength{\abovedisplayskip}{5pt}
		\setlength{\belowdisplayskip}{5pt}
		\hat{h}_{n,c,y,x}=\alpha_{c,y,x}\frac{h_{n,c,y,x}-\mu_c}{\sigma_c}+\beta_{c,y,x},
		\end{equation}
		where $\mu_c$ and $\sigma_c$ are the mean and standard deviation of $h$ in the channel $c$. $\alpha$ and $\beta$ are the weighted sum of these features from the convolution output of the target semantic map $\hat{S}_{p_t}$ and the target style map $\mathcal{C}$, which are defined as:
		\begin{equation}
		\setlength{\abovedisplayskip}{5pt}
		\setlength{\belowdisplayskip}{5pt}
		\begin{aligned}
		\alpha = \theta_\alpha \cdot h_\alpha^s + (1-\theta_\alpha) \cdot h_\alpha^c\\
		\beta = \theta_\beta \cdot h_\beta^s + (1-\theta_\beta) \cdot h_\beta^c
		\end{aligned}
		\end{equation}
		where $\theta_\alpha$ and $\theta_\beta$ are the learnable parameters. It should be noted that the target style map $\mathcal{C}$ is shared in all the SPGBlock and the parameters in the style map generation are optimized by the gradient from the final objective $\mathcal{L}_{full}$.
	}
	
	{Finally, after several stacks of the UpBlock and the {SPGBlock} on different feature scales, we can generate the target results $\hat{I}_{p_t}$ in a coarse-to-fine manner.
	}
	
	\subsection{Learning Objective}
	{To train the proposed SPATN network, which is suggested in the first stage to predict the target semantic map $\hat{S}_{p_t}$, we adopt a cross-entropy loss to constrain the predicted $\hat{S}_{p_t}$ close to its ground-truth label $S_{p_t}$, which is defined as:}
	\begin{equation}
	\setlength{\abovedisplayskip}{4pt}
	\setlength{\belowdisplayskip}{4pt}
	\mathcal{L}_{ce}=-\sum_{i, j}\sum_c{S_{p_t}(i, j, c)}log(\hat{S}_{p_t}({i, j, c})),
	\end{equation}
	{where $i$, $j$ and $c$ denote the positions of each semantic maps.}

	{For the second stage, our proposed SPGNet is trained to generate a photo-realistic person image with the target pose $p_t$. 
		The predicted $\hat{I}_{p_t}$ is expected to approximate the ground-truth image $I_{p_t}$ in both pixel and perceptual levels, and more importantly, should follow the real person image distribution. Thus, the learning objective to train the SPGNet contains two parts, \ie, reconstruction loss and photo-realistic loss.}
	
	\noindent\textbf{Reconstruction Loss.} {In general, it contains two terms, \ie, L1 loss $\mathcal{L}_{L1}$ and perceptual loss $\mathcal{L}_{perc}$~\cite{johnson2016perceptual}. The L1 loss is defined on the pixel space to measure the appearance difference between the generated results $\hat{I}_{p_t}$ and the ground-truth $I_{p_t}$, which is defined as:}
	\begin{equation}
	\setlength{\abovedisplayskip}{4pt}
	\setlength{\belowdisplayskip}{4pt}
	\mathcal{L}_{L1} = \frac{1}{CHW}||\hat{I}_{p_t}-I_{p_t}||_1,
	\end{equation}
	\vspace{-2pt}
	{where $C$, $H$ and $W$ represent the image dimensions.The perceptual loss $\mathcal{L}_{perc}$ is defined on the feature space, which is usually adopted to improve the visual quality of the generated images. In particular, we adopt the pre-trained VGG19 model~\cite{simonyan2014very} to extract the features in multi-scale space, which is defined as:}
	\begin{equation}
	\setlength{\abovedisplayskip}{4pt}
	\setlength{\belowdisplayskip}{4pt}
	\mathcal{L}_{perc}=\sum_{k=1}^K\frac{1}{C_kH_kW_k}||\phi_k(\hat{I}_{p_t})-\phi_k(I_{p_t})||_2^2,
	\end{equation}
	\vspace{-2pt}
	{where $\phi_k$ denotes the $k$-\textit{th} layer of the pre-trained VGG19 model $\phi$. In this work, we set $K=4$.} 
	
	\noindent\textbf{Photo-realistic Loss.} 
	{Adversarial loss~\cite{goodfellow2014generative} has been proved to be very effective in generating realistic results. Therefore, we also adopt it in our work to constrain the generated results $\hat{I}_{p_t}$ to be more photo-realistic. In this work, we take the triplet $\{I_{p_t}, I_{p_s}, p_t\}$ and $\{\hat{I}_{p_t}, I_{p_s}, p_t\}$ as the input of discriminator $D$ to constrain the generated result $\hat{I}_{p_t}$ to have the same appearance with $I_{p_s}$ and the same pose with $p_t$. In this work, we adopt the PatchGAN~\cite{isola2017image} to judge the realism of image patches. The objective for training the discriminator $D$ and generator $G$ (SPGNet in this work) is defined as:}
	\begin{equation}
	\setlength{\abovedisplayskip}{4pt}
	\setlength{\belowdisplayskip}{4pt}
	\begin{aligned}
	\mathcal{L}_{adv}(G,D)=&\mathbb{E}_{I_{p_s},I_{p_t}}[\log(1-D(G(p_t,I_{p_s},\hat{S}_{p_t} )|I_{p_s},p_t))]
	\\+&\mathbb{E}_{I_{p_s},I_{p_t}}[\log D(I_{p_t}|I_{p_s},p_t)].
	\end{aligned}
	\end{equation}
	\vspace{-2pt}
	{By taking the reconstruction loss and photo-realistic loss into consideration, the overall learning objective of our framework including SPATN and SPGNet is formulated as:}
	\begin{equation}
	\setlength{\abovedisplayskip}{4pt}
	\setlength{\belowdisplayskip}{4pt}
	\mathcal{L}_{full}=\lambda_{ce}\mathcal{L}_{ce}+\lambda_{L1}\mathcal{L}_{L1}+\lambda_{perc}\mathcal{L}_{perc} + \lambda_{adv}\mathcal{L}_{adv},
	\end{equation}
	{where $\lambda_{ce}$, $\lambda_{L1}$, $\lambda_{perc}$, $\lambda_{adv}$ are the trade-off parameters.
	}
	\section{Experiments}
	{Extensive experiments are conducted to assess the effectiveness of our SPGNet, and compare it with the recent state of the art methods, including Def-GAN~\cite{siarohin2018deformable}, Intr-Flow~\cite{li2019dense}, PATN~\cite{zhu2019progressive}, GFLA~\cite{ren2020deep}, ADGAN~\cite{men2020controllable}, {XingGAN~\cite{tang2020xinggan}, and BiGraphGAN~\cite{tang2020bipartite}}. Quantitative and qualitative {results} as well as user study are reported for a comprehensive comparison. In addition, we also conduct an ablation study to explore the benefits that our method may bring to the final results. More results can be found in our supplemental materials.} 
	\subsection{Dataset and Experimental Details}
	\noindent\textbf{Dataset.} {There are two datasets that are usually used in this task, \ie, DeepFashion (In-shop Clothes Retrieval Benchmark)~\cite{liu2016deepfashion} and Market-1501~\cite{zheng2015scalable}. As for the first one, it contains a total of 52,712 person images with clean background and resolution of $256\times256$, which cover various poses and appearances. Following the same settings in \cite{zhu2019progressive}, 
		{we divide this dataset into training and testing subsets with 101,966 and 8570 pairs, respectively.}
		In contrast, Market-1501~\cite{zheng2015scalable} is more challenging due to its relatively low resolution ($128\times64$), complex background, diverse light conditions, etc. Here, we follow the data split as \cite{zhu2019progressive}, 
		{in which 263,632 pairs are selected for training and 12,000 ones for testing.}
		These {two} subsets are not overlapped in terms of either identity and image. For building the body structure, we adopt OpenPose~\cite{cao2017realtime} to extract the 18 human joints. Since our SPATN model (the 1-\textit{st} stage) is designed to take the source parsing map $S_{p_s}$ to predict the target one $S_{p_t}$, it is necessary to obtain the parsing maps $S_{p_s}$ and $S_{p_t}$ for training. For the training data of DeepFashion~\cite{liu2016deepfashion}, the off-the-shelf PGN model~\cite{gong2018instance} is utilized to predict the human parsing maps by the given human images, which are regarded as the ground-truth for training the SPATN model. However, due to the adverse effect of low quality and complex background, PGN~\cite{gong2018instance} fails to generate the plausible {parsing maps with the given human images} from Market-1501, making it unable to predict the target parsing maps, which inevitably result in the failure of the latter per-region translation. To tackle this problem, we train another SPATN* model on DeepFashion by using the source appearance image (rather than the source parsing map) and target pose to predict the target parsing map of Market-1501. This can eliminate the dependency of source parsing maps and achieve superior performance on predicting the target parsing maps for Market-1501.}

	\noindent\textbf{Implementation Details.} {We adopt the Adam optimizer~\cite{kingma2014adam} with $\beta_1=0.5, \beta_2=0.999$ in all experiments. The learning rate $lr$ is set to $2\times 10^{-4}$ for the {two-stage model} and $2\times 10^{-5}$ for the discriminator, respectively. The $lr$ is decreased by 0.5 when the reconstruction loss on the validation set is no longer decreasing. The trade-off parameters are set as follows: $\lambda_{ce}=10$, $\lambda_{L1}=1$, $\lambda_{perc}=1$, $\lambda_{adv}=0.01$. The batch size is set to 4 and 32 for training on DeepFashion and Market-1501, respectively. 
		{The SPATN and SPGNet are trained simultaneously, in which the SPGNet takes the ground-truth target parsing map $S_{p_t}$ as input. Only in the inference, SPGNet takes the predicted target parsing map $\hat{S}_{p_t}$ from SPATN model as input.}
		up{More details and analyses about the training manner are given in the suppl.}
		The whole model and experiments are carried out on a PC server with 2 1080Ti. It takes one week to train the {two-stage model}.}

	\noindent\textbf{Evaluation Metrics.} {Though it remains an open problem to quantitatively evaluate the quality of the appearance and shape consistency of the generated results, here we adopt four metrics, \ie, SSIM, FID, LPIPS, and PCKh, which are common used in the previous works. Among these metrics, SSIM~\cite{wang2004image} is proposed to evaluate the quality of generated images by computing the global variance and means of the image to assess the structure similarity. Fréchet Inception Distance (FID)\cite{heusel2017gans} is adopted to measure the realism of the generated images by calculating the Wasserstein-2 distance between the distributions of the generated results and its corresponding ground-truth. LPIPS~\cite{zhang2018unreasonable} is another common metric to assess the visual quality, which is more consistent with human perception. PCKh~\cite{andriluka20142d} is recently suggested to measure the shape consistency, in which the score is computed by measuring the accuracy of the localization of the body joints. In addition, user study is conducted to evaluate the visual quality and the faithfulness (\ie, realistic appearance and consistent shape). Except these quantitative metrics, we report the visual results to compare with these competing methods to show the superiority of our SPGNet. }

	\subsection{Quantitative and Qualitative Comparison}
	
	\begin{table}[t]
		\vspace{4pt}
		\begin{center}
			\small
			\setlength{\tabcolsep}{1.8mm}
			{
				\begin{tabular}{|c|c|c|c|c|}
					\hline
					\rowcolor{lightgray}
					\textbf{Method} & \textbf{SSIM $\uparrow$} & \textbf{FID $\downarrow$} & \textbf{PCKh $\uparrow$} & \textbf{LPIPS $\downarrow$} \\
					\hline
					Def-GAN\cite{siarohin2018deformable} & 0.760 & 18.475 & 0.94 & 0.2330 \\
					PATN\cite{zhu2019progressive} & 0.773 & 20.739 & 0.96 & 0.2533 \\
					Intr-Flow\cite{li2019dense} & 0.778 & 16.314 & 0.96 & 0.2131\\
					GFLA\cite{ren2020deep} & \textbf{0.790} & \textbf{10.573} & 0.96 & 0.2341  \\
					ADGAN\cite{men2020controllable} & 0.772 & 14.460 &0.96 &0.2256 \\
					XingGAN\cite{tang2020xinggan} & 0.778 & 39.322 & 0.95 & 0.2927 \\
					BiGraphGAN\cite{tang2020bipartite} & 0.778 & 20.951 & \textbf{0.97} &0.2444 \\
					Ours & 0.782 & 12.243 & \textbf{0.97} & \textbf{0.2105} \\
					\hline
				\end{tabular} 
			}
		\end{center}
		\vspace{-6pt}
		\caption{Quantitative comparisons on two DeepFashion test sets.}
		\vspace{-18pt}
		\label{tab:deepf}
	\end{table}
	\begin{table*}[t]
		\vspace{-15pt}
		\begin{center}
			\scriptsize
			\renewcommand\arraystretch{1.2}
			\setlength{\tabcolsep}{1.26mm}
			{
				\begin{tabular}{|c|c|c|c|c|c|c|c|c|c|c|c|c|}
					\hline
					\rowcolor{lightgray}
					& \multicolumn{6}{c|}{\textbf{Comparison on Market-1501}} & \multicolumn{6}{c|}{\textbf{User Study}}\\
					\hhline{>{\arrayrulecolor{lightgray}}-|>{\arrayrulecolor{black}}------------}
					\rowcolor{lightgray} \multirow{-2}{*}{\makecell[c]{\textbf{Methods}}} & \textbf{SSIM $\uparrow$} & \textbf{M-SSIM $\uparrow$} & \textbf{FID $\downarrow$} & \textbf{PCKh $\uparrow$} & \textbf{LPIPS $\downarrow$}  & \textbf{M-LPIPS$\downarrow$} & \textbf{R2G$\uparrow$ (DF)} & \textbf{G2R$\uparrow$ (DF)} & \textbf{R2G$\uparrow$ (M)}  & \textbf{G2R$\uparrow$ (M)} & \textbf{Jab$\uparrow$ (DF)} &\textbf{Jab$\uparrow$ (M)}\\
					\hline
					Def-GAN\cite{siarohin2018deformable}         & 0.290 & 0.805 & 25.364 & 0.94 & 0.2994 & 0.1496 & 12.42 & 24.61 & 22.67 & 50.24& 4.87 & 11.07\\
					PATN\cite{zhu2019progressive}            & 0.311 & 0.811 & 22.657 & 0.94 & 0.3196 & 0.1590 & 19.14 & 31.78 & 32.23 & 63.47 &8.27 & 6.53 \\
					Intr-Flow\cite{li2019dense}       & 0.308   & 0.813  & 27.163  & 0.95  & 0.2888  & 0.1403  & 10.01 & 31.71 & 36.27 &  65.33 &13.60 &17.07\\
					GFLA\cite{ren2020deep}            & 0.281 & 0.796 &  \textbf{19.751} & 0.94 & 0.2817 & 0.1482& 19.53 & 35.07 & 35.87 & 64.93 & 22.60 & 16.80 \\
					XingGAN\cite{tang2020xinggan}         & 0.313 & 0.816 & 22.495 & 0.93 & 0.3059 & 0.1581& 21.61 & 33.75 & 35.28 & 65.16 & 4.73 &7.20 \\
					BiGraphGAN\cite{tang2020bipartite} & \textbf{0.325} & \textbf{0.818} & 28.915 &0.94 &0.3048 & 0.1505& \textbf{22.39} &34.16 &35.76 &65.91 & 13.93 &16.20 \\
					Ours & 0.315 &  \textbf{0.818} & 23.331 & \textbf{0.97} & \textbf{0.2779} & \textbf{0.1385} & 19.47 & \textbf{36.80} & \textbf{37.93} & \textbf{66.53} & \textbf{32.00} & \textbf{25.13} \\
					\hline
				\end{tabular}
			}
		\end{center}
		\vspace{-4pt}
		\caption{The quantitative comparison with the competing methods on Market-1501 test set and the two groups comparisons of user study. Here, DF (M) represents the evaluation on DeepFashion (Market-1501) test datasets. $\uparrow$ ($\downarrow$) indicates higher (lower) is better.}
		\label{tab:market}
		\vspace{-4pt}
	\end{table*}
	\begin{figure*}[!t]
		\vspace{1pt}
		\centering
		\includegraphics[width= 1.0\linewidth]{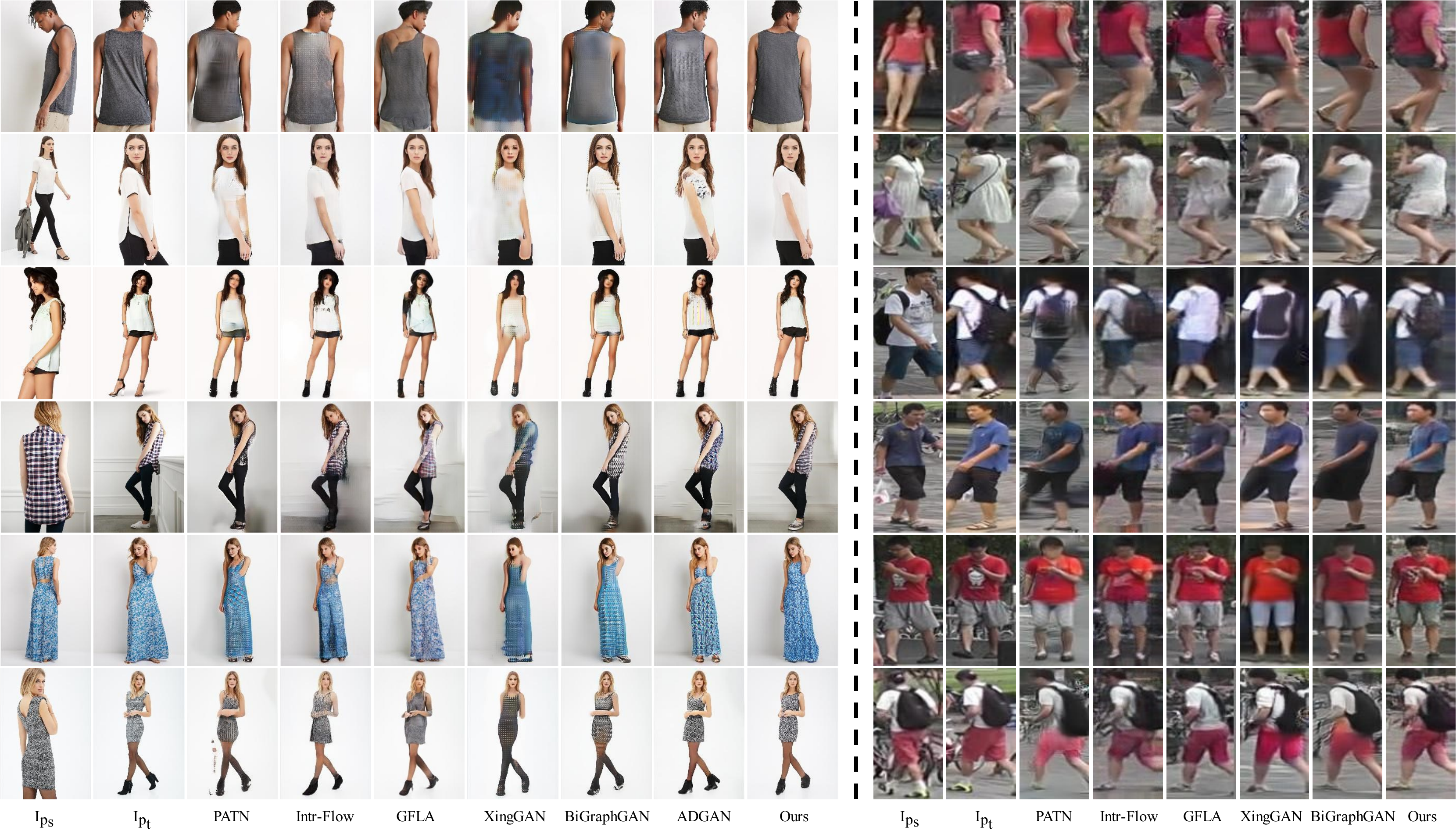}
		\caption{{The visual comparison with the competing methods on two types of test sets. Best view it by zooming in the screen.}}
		\label{fig:comparison}
		\vspace{-2pt}
	\end{figure*}

	{We compare our proposed model with several recent state of the art methods, including Def-GAN~\cite{siarohin2018deformable}, Intr-Flow~\cite{li2019dense}, PATN~\cite{zhu2019progressive}, GFLA~\cite{ren2020deep},  XingGAN~\cite{tang2020xinggan}, {and BiGraphGAN~\cite{tang2020bipartite}}. The quantitative result on DeepFashion is shown in Table~\ref{tab:deepf}. We can observe that our proposed method achieves comparable performance on this dataset, which means that results of ours own high visual quality and are {realistic}. In addition, we also conduct the evaluation on the challenging dataset Market-1501. The comparison is shown in {the left part of} Table~\ref{tab:market}. Mask-SSIM {(M-SSIM)} and Mask-LPIPS {(M-LPIPS)} are also considered to exclude the effect of irrelevant regions, \ie, background. We can see that our method can also have a superior performance to these competing methods, indicating the great generalization of our method. The superior performance can be attributed to the proposed two-stage model in which the region adaptive normalization is utilized for per-region style translation.}
	
	{Figure~\ref{fig:comparison} gives the qualitative comparison on both DeepFashion and Market-1501 datasets. We can have the following observations, (i) though the source image $I_{p_s}$ is under an extreme pose (1-\textit{st} and 3-\textit{rd} rows in the left part), results of our method are more realistic, and especially, have the consistent shape with the ground-truth $I_{p_t}$, indicating the benefits of the SPATN that bring to the generation process. (ii) In terms of complex texture (4$\sim$6-\textit{th} rows in the left part), ours are {obviously} superior to the competing methods in retaining the appearance textures, which is mainly attributed to the utilization of per-region style translation. (3) Within the complex background, the results of ours on the Market-1501 (right part of Figure~\ref{fig:comparison}) still {perform favorably} in generating the clear and  photo-realistic results as well as a consistent appearance with the source image.}

	\subsection{User Study}
	Although the above evaluation indicators can evaluate the performance of the generated results from different aspects, the person image generation task is likely to be user-oriented. {Therefore, we conduct a user study on the two test sets to evaluate the performance from real human perception. To be specific, it mainly contains two groups that are conducted from different aspects. (i) Comparison with ground-truth. Following~\cite{ma2017pose,siarohin2018deformable}, we randomly select 55 real and 55 generated images from the test set and then shuffle them. The first 10 of the 110 images are used for practice and the remaining 100 images are used to assess the performance. 30 volunteers that cover the bachelor and master students with computer vision background are required to give a choice about whether this one is real or generated within a second. (ii) Comparison with the state of the arts. To achieve this goal, we randomly select 55 image pairs, including source image, target pose, ground-truth and images generated by these competing methods. Volunteers are required to select the image that is the closest to the ground-truth in terms of the visual quality, and more importantly, the pose and appearance consistency with ground-truth. Results are shown in the right part of Table~\ref{tab:market}. Here we adopt three metrics, \ie, {R2G}: the percentage of the real image that is judged as the generated one, {G2R}: the percentage of the generated image that is judged as real one, {Jab}: the percentage that the image is judged as the best one. Higher values in these three metrics indicate better performance. We can see that results of ours outperform the state of the arts with a large margin (\eg, 9.40\% and 8.06\% higher than the 2-\textit{nd} best one in Jab), which further indicates the superiority of our method.}

	\subsection{Ablation Study}
	\begin{figure}[!t]
		\centering
		\includegraphics[width=1\linewidth]{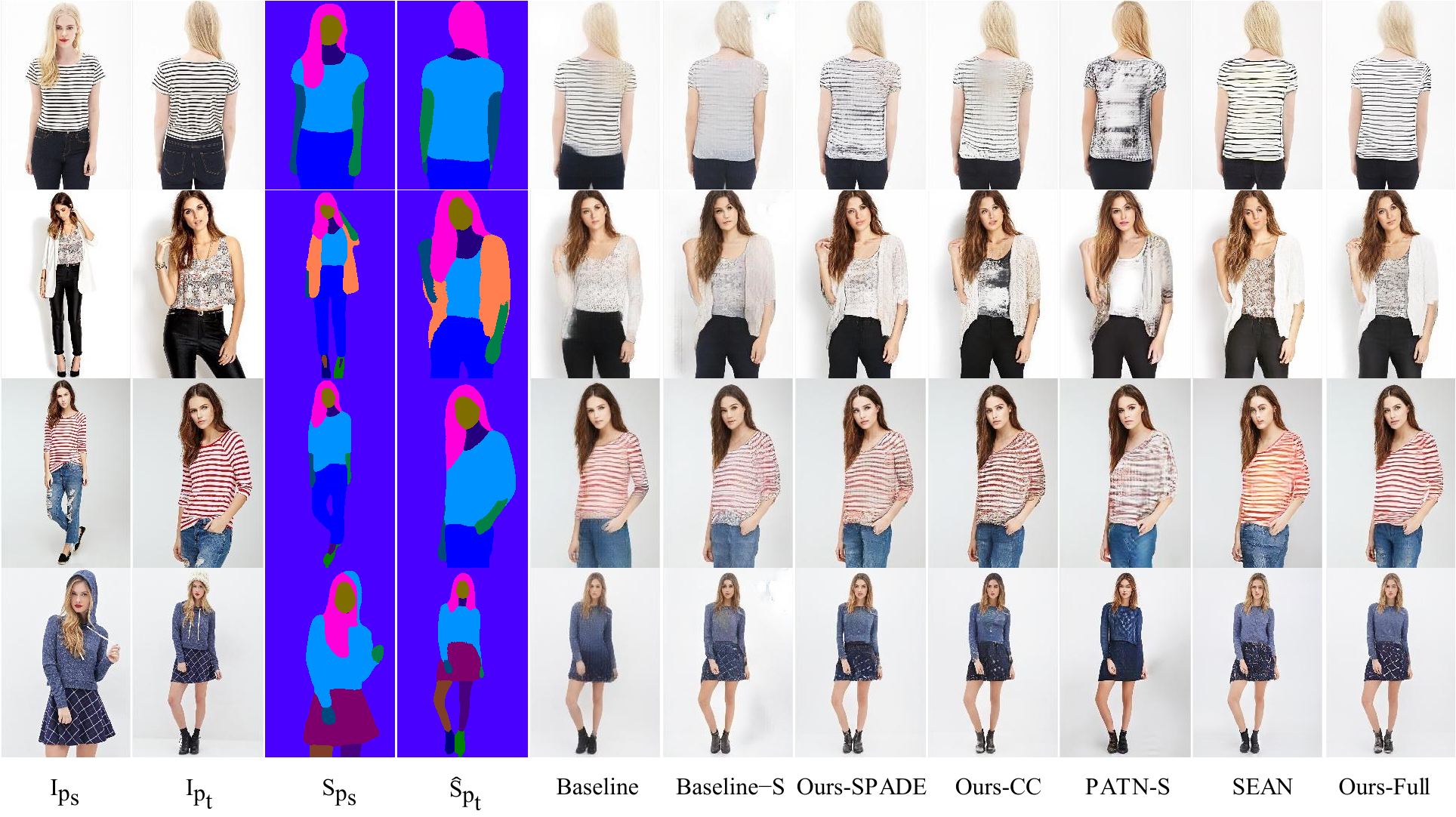}
		\vspace{-8pt}
		\caption{{Visual comparison of different SPGNet variants}.}
		\label{fig:ablation}
		\vspace{-16pt}
	\end{figure}

	up{In order to explore the effectiveness of our per-region style translation in the human generation task, in this paper, we follow \cite{li2019dense} and conduct the ablation study on the DeepFashion datasets with the following variants, 
		(i) Baseline: our baseline model is a dual-path U-Net, which mainly comes from Intr-Flow\cite{li2019dense}. There are two encoders, which encode appearance and pose, respectively. The decoder combines the target pose and the warped appearance features to generate the final image. It should be noted that there is not any conditional normalization in this model. 
		(ii) Baseline-S: based on the Baseline model, the pose encoder of Baseline-S takes the additional features (\ie, the target semantic parsing map that is predicted by our SPATN model) as input.
		(iii) Ours-SPADE: by replacing the per-region adaptive instance normalization (SEAN) with the spatial adaptive normalization (SPADE)~\cite{park2019semantic}. 
		(iv)  Ours-CC~\cite{liu2019learning}: by replacing SEAN with CC-FPSE in the decoder module, in which ours-CC takes the target semantic parsing map as input and predicts the conditional convolution kernels to guide image generation.
		(v) PATN-S: by taking the semantic parsing maps as additional input data to retrain PATN. 
		(vi) SEAN: by using the SEAN generator~\cite{zhu2020sean} to directly broadcast the style information to the target semantic layout to generate an image. 
		(vii) Ours-Full. 
		The quantitative results are shown in Table~\ref{tab:ablation}. We can see that 
		(1) compared with Baseline and Baseline-S, PATN and PATN-S, though slight improvements can be obtained, Ours-Full obviously outperforms them with a large margin, especially the FID metric (\eg, Baseline-S is 13.1\% higher than Baseline), indicating the benefits that the per-region adaptive normalization brings to this task.
		(2) Compared with the Baseline model, Ours-SPADE and Ours-CC achieve an obvious improvement by incorporating the spatial adaptive normalization and the conditional convolution, respectively, but they are still inferior to Ours-Full. 
		(3) Directly taking SEAN to handle this task, the poor performance may be caused by the inconsistent poses, indicating the necessity of feature deformation in Ours-Full.
		(4) With the per-region adaptive normalization, Ours-Full performs superior to others, indicating the effectiveness of our SPGNet in human person generation. 
		The visual comparison is shown in Fig.~\ref{fig:ablation}. We can see that compared with other variants, Ours-Full obviously outperforms others in generating photo-realistic and consistent appearance results, {which further indicates the benefits of our SPGNet by adopting the per-region style translation in handling this task}.}
	\begin{table}
		\begin{center}
			\small
			\renewcommand\arraystretch{1.1}
			\setlength{\tabcolsep}{2.51mm}
			{
				\begin{tabular}{|c|c|c|c|c|}
					\rowcolor{lightgray}
					\hline
					\textbf{Variants} & \textbf{SSIM $\uparrow$} & \textbf{FID $\downarrow$} & \textbf{PCKh $\uparrow$} & \textbf{LPIPS $\downarrow$} \\
					\hline
					Baseline & \textbf{0.784} & 18.946 & 0.96 & 0.2308 \\
					Baseline-S &0.780 & 16.460 & 0.96 & 0.2353 \\
					Ours-SPADE & 0.779 & 13.808 & 0.96 & 0.2200 \\
					Ours-CC &0.778 &13.686 &0.96 &0.2252 \\
					PATN-S &0.776& 15.488 &0.96 & 0.2557 \\
					SEAN &0.778 & 15.993 &0.96 & 0.2209\\
					Ours-Full& 0.783 & \textbf{12.613}  & \textbf{0.97} &\textbf{0.2133} \\
					\hline
				\end{tabular}
			}
		\end{center}
		\vspace{-4pt}
		\caption{Quantitative comparison of different SPGNet variants.}
		\label{tab:ablation}
		\vspace{-14pt}
	\end{table}
	
	\section{Conclusion}

	up{In this paper, we propose a two-stage model to deal with the challenging pose transfer task. In the first stage, we generate the target semantic parsing maps to eliminate the difficulties of pose transfer and benefit the latter per-region appearance translation. In the second one, with the predicted semantic map, we adopt the region adaptive normalization to achieve the per-region style translation, which is more effective in the person image generation task. The proposed method decomposes the complex task into two easier problems. 
		Experiments show that both per-region style and semantic map are crucial in generating high-quality body parts and fine-scale textures. Moreover, instead of joint heatmaps, the distance map of the skeleton is adopted for a better generation of target semantic parsing map.
		Both the quantitative and visual comparison demonstrate the superior performance in generating consistent appearance and photo-realistic results with complex poses and source appearance. }

	\noindent\textbf{Acknowledgments.} {This work was supported in part by the National Natural Science Foundation of China under grant No. U19A2073.}
	\clearpage
	{\small
		\bibliographystyle{ieee_fullname}
		\bibliography{egbib}
	}
	\clearpage
	
	\clearpage
	\section*{Appendix}
	\subsection*{A. Training procedures}
	{In order to explore the training manners of the two-stage model that may bring benefits to the final results, in this supplemental material, we conduct three schemes, (i) $S_1 \triangleright S_2$: pre-training the SPATN model in stage one, then fixing its parameters and training SPGNet in stage two; (ii) $S_1\! \Join \!S_2$: pre-training the SPATN model in stage one, then jointly training with the SPGNet in stage two; (iii) $S_1\!\parallel \!S_2$: training the SPATN and SPGNet simultaneously, in which the SPGNet takes the ground-truth target parsing map ${S}_{p_t}$ as input. During the inference, the model takes the predicted target parsing map $\hat{S}_{p_t}$ from the SPATN model as input to generate the final results. The comparison results are shown in Table~\ref{tab:train}. We find that though the joint training manner ($S_1\! \Join \!S_2$) usually obtains great performance in many fields, it shows inferior results in this task. We analyze that 1) due to the long pathway of SPGNet, the gradient from the $\mathcal{L}_{full}$ to update the SPATN model tends to be zero, making the SPATN model seldomly benefits from the end-to-end training, 2) more importantly, there is a gap between the predicted semantic parsing map and the ground-truth. The inaccurate prediction may confuse the learning of the SPGNet, thus has a negative impact on the generation process. Therefore, we adopt the $S_1\!\parallel \!S_2$ scheme to train our two-stage model and use the predicted target semantic maps in the first stage to guide the final generation of the target image in the inference. }
	\vspace{-10pt}
	\begin{table}[H]
		\begin{center}
			\small
			\setlength{\tabcolsep}{3.1mm}
			{
				\begin{tabular}{|c|c|c|c|c|}
					\hline
					\rowcolor{lightgray}
					\textbf{Scheme}& \textbf{SSIM $\uparrow$} & \textbf{FID $\downarrow$} & \textbf{PCKh $\uparrow$} & \textbf{LPIPS $\downarrow$} \\
					\hline
					$S_1 \triangleright S_2$ & 0.785 & 17.340 & 0.96 &  0.2240 \\
					$S_1\! \Join \!S_2$ & \textbf{0.792} & 24.309 & 0.96 &  0.2435 \\
					$S_1\parallel S_2$ & 0.782 & \textbf{12.243} & \textbf{0.97} &  \textbf{0.2105} \\
					\hline
				\end{tabular}
			}
		\end{center}
		\vspace{-5pt}
		\caption{The quantitative comparison of different training schemes on our DeepFashion test set. $\uparrow$ ($\downarrow$) means higher (lower) is better.}
		\label{tab:train}
	\end{table}
	\vspace{-15pt}
	
	\subsection*{B. Network Architecture of SPGNet}
	Our SPGNet consists of a pose encoder, an appearance encoder and a decoder which is composed of several SPGBlocks. Table \ref{tab:SPGNet} shows the details of SPGNet. Table \ref{tab:SPGBlock} and Table \ref{tab:FeatureWarp} give the detailed network architecture of SPGBlock and Feature Deformation Module, respectively. Conv. ($d,k,s$) and ConvT.($d,k,s$) denote convolution and transposed convolution layer, where $d$, $k$ and $s$ are output dimention, convolution kernel size and stride, respectively. And $dim({f_{t-1}})$ is the dimention of feature map $f_{t-1}$. LReLU is leaky ReLU with negative slope $c$. BN and IN represent batch normalization and instance normalization, respectively. The denotation of input is the same as that in the paper. $I_{p_s}$ denotes the source appearance image. $S_{p_s}$ and $\hat{S}_{p_t}$ are source semantic parsing map and predicted target semantic parsing map. $p$ is keypoint heat map of the target pose. $\Phi_{2D}$ and $V$ denote the projection of the predicted 3D flow and visibility map from Intr-Flow~\cite{li2019dense}, respectively.

	\begin{table*} 
		\begin{center}
			\small
			\renewcommand\arraystretch{1.1}
			\setlength{\tabcolsep}{2.52mm}
			{
				\begin{tabular}{|c|c|c|c|}
					\hline
					\textbf{Input} & 
					\makecell{$I_{p_s}$ ($3\times\times256\times256$)\\
						$\Phi_{2D}$ ($2\times\times256\times256$)\\
						$V$ ($1\times\times256\times256$) \\
					}&
					\makecell{$p$ \\ ($18\times256\times256$)} & 
					\makecell{$I_{p_s}$, ($3\times256\times256$) \\$S_{p_s}$, $\hat{S}_{p_t}$, ($20\times256\times256$)}\\
					\hline
					\textbf{\makecell{Feature \\ Extraction}}& 
					\makecell{
						Conv. (32, 1, 1)\\
						ResidualBlock, ResidualBlock, FeatureWarp\\
						ReLU, Conv. (64, 3, 2), BN \\
						ResidualBlock, ResidualBlock, FeatureWarp\\
						ReLU, Conv. (128, 3, 2), BN \\
						ResidualBlock, ResidualBlock, FeatureWarp\\
						ReLU, Conv. (128, 3, 2), BN \\
						ResidualBlock, ResidualBlock, FeatureWarp\\
						ReLU, Conv. (128, 3, 2), BN \\
						ResidualBlock, ResidualBlock, FeatureWarp\\
						ReLU, Conv. (128, 3, 2), BN \\
						ResidualBlock, ResidualBlock\\
						ReLU, Conv. (128, 3, 2), BN \\
						ResidualBlock, ResidualBlock\\
						ReLU, Conv. (128, 3, 2), BN
					}
					& 
					\makecell{
						Conv. (32, 1, 1)\\
						ResidualBlock, ResidualBlock\\
						ReLU, Conv. (64, 3, 2), BN \\
						ResidualBlock, ResidualBlock\\
						ReLU, Conv. (128, 3, 2), BN \\
						ResidualBlock, ResidualBlock\\
						ReLU, Conv. (128, 3, 2), BN \\
						ResidualBlock, ResidualBlock\\
						ReLU, Conv. (128, 3, 2), BN \\
						ResidualBlock, ResidualBlock\\
						ReLU, Conv. (128, 3, 2), BN \\
						ResidualBlock, ResidualBlock\\
						ReLU, Conv. (128, 3, 2), BN \\
						ResidualBlock, ResidualBlock\\
						ReLU, Conv. (128, 3, 2), BN
					}
					&
					\makecell{
						Conv. (32, 3, 1), IN, LReLU(0.2)\\
						Conv. (64, 3, 2), IN, LReLU(0.2)\\
						Conv. (128, 3, 2), IN, LReLU(0.2)\\
						ConvT.(64, 3, 2), IN, LReLU(0.2)\\
						ConvT.(32, 3, 2), IN, LReLU(0.2)\\
						ConvT.(128, 3, 1), Tanh \\
						Region Average Pooling 
					}
					\\
					\hline
					\textbf{output}&
					\makecell{
						warped appearance features $f_a^w$
					} &
					\makecell{
						pose features $f_p$
					} &
					\makecell{
						Style Codes $ST$
					} \\
					\hline
					\textbf{\makecell{Feature \\ Fusion}} &
					\multicolumn{3}{c|}{
						\makecell{
							ReLU, Conv. (512, 3, 1) PixelShuffle(2), SPGBlock 1\\
							ReLU, Conv. (512, 3, 1) PixelShuffle(2), SPGBlock 2 \\
							ReLU, Conv. (512, 3, 1) PixelShuffle(2), SPGBlock 3 \\
							ReLU, Conv. (512, 3, 1) PixelShuffle(2), SPGBlock 4 \\
							ReLU, Conv. (384, 3, 1) PixelShuffle(2), SPGBlock 5 \\
							ReLU, Conv. (256, 3, 1) PixelShuffle(2), SPGBlock 6 \\
							ReLU, Conv. (128, 3, 1) PixelShuffle(2), SPGBlock 7 \\
							Conv. (3, 7, 1) \\
							Tanh()
						}
					}\\
					\hline
					\textbf{output}& \multicolumn{3}{c|}{\makecell{$\hat{I}_{p_t}$\\($3\times256\times256$)}} \\
					\hline
				\end{tabular}
			}
		\end{center}
		\caption{Details of SPGNet Architecture.}
		\vspace{-2pt}
		\label{tab:SPGNet}
	\end{table*}
	
	\begin{table} 
		\begin{center}
			\small
			\renewcommand\arraystretch{1.1}
			\setlength{\tabcolsep}{4mm}
			{
				\begin{tabular}{|c|c|c|c|c|c|}
					\hline
					\textbf{Input} & $f_{t-1}$ & $f_a^w$ & $f_p$ & $ST$ & $\hat{S}_{p_t}$\\
					\hline
					
					\multirow{7}{*}{\textbf{SPGBlock}}
					& $f_{t-1}$ & \multicolumn{2}{c|}{Concat} & \multicolumn{2}{c|}{broadcasting} \\\cline{2-6}
					
					& $f_{t-1}$ & \multicolumn{4}{c|}{\makecell{SEAN, ReLU\\ Conv. ($dim({f_{t-1}})$, 1, 1)}}      \\\cline{2-6}
					& \multicolumn{5}{c|}{Concat}           \\\cline{2-6}
					& \multicolumn{5}{c|}{SEAN, ReLU}        \\\cline{2-6}
					& \multicolumn{5}{c|}{ Conv. ($dim(f_{t-1})$, 3, 1)}  \\\cline{2-6}
					& \multicolumn{5}{c|}{$+f_{t-1}$} \\
					\hline
					\textbf{output}& \multicolumn{5}{c|}{$f_t$} \\
					\hline
				\end{tabular}
			}
		\end{center}
		\caption{Details of SPGBlock.}
		\vspace{-4pt}
		\label{tab:SPGBlock}
	\end{table}
	
	\begin{table} 
		\begin{center}
			\small
			\renewcommand\arraystretch{1.1}
			\setlength{\tabcolsep}{4.8mm}
			{
				\begin{tabular}{|c|c|c|c|}
					\hline
					\textbf{Input} & $f_{a}$ & $\Phi_{2D}$ & $V$  \\
					\hline
					\multirow{7}{*}{\makecell[c]{\textbf{Feature}\\\textbf{Deformation}}} &
					\multicolumn{2}{c|}{Warp} & $V$ \\\cline{2-4}
					& \multicolumn{3}{c|}{\makecell{Expand Feature\\ $f_a^{w'}*(V==1)$\\$f_a^{w'}*(V==0)$}} \\\cline{2-4}
					& \multicolumn{2}{c|}{Visible part}&\multicolumn{1}{c|}{Invisible part} \\\cline{2-4}
					& \multicolumn{3}{c|}{Concat}\\\cline{2-4}
					& \multicolumn{3}{c|}{ResidualBlock}\\
					\hline
					\textbf{output}& \multicolumn{3}{c|}{$f_a^w$} \\
					\hline
				\end{tabular}
			}
		\end{center}
		\caption{Details of Feature Deformation Module.}
		\vspace{-4pt}
		\label{tab:FeatureWarp}
	\end{table}
	
	\vspace{-2pt}
	\subsection*{C. Distance Map}
	\begin{figure}[h]
		\centering
		\includegraphics[width=1.\linewidth]{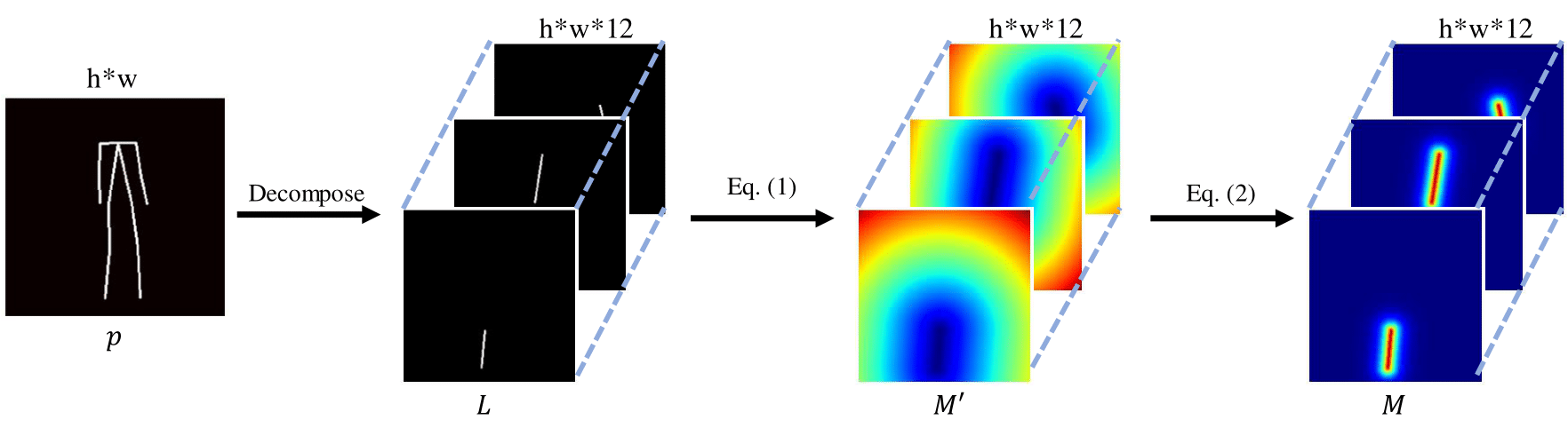}
		\caption{The details of the distance map generation.}
		\label{fig:distance map}
	\end{figure}
\begin{figure*}[!t]
		\centering
		\includegraphics[width=.62\linewidth]{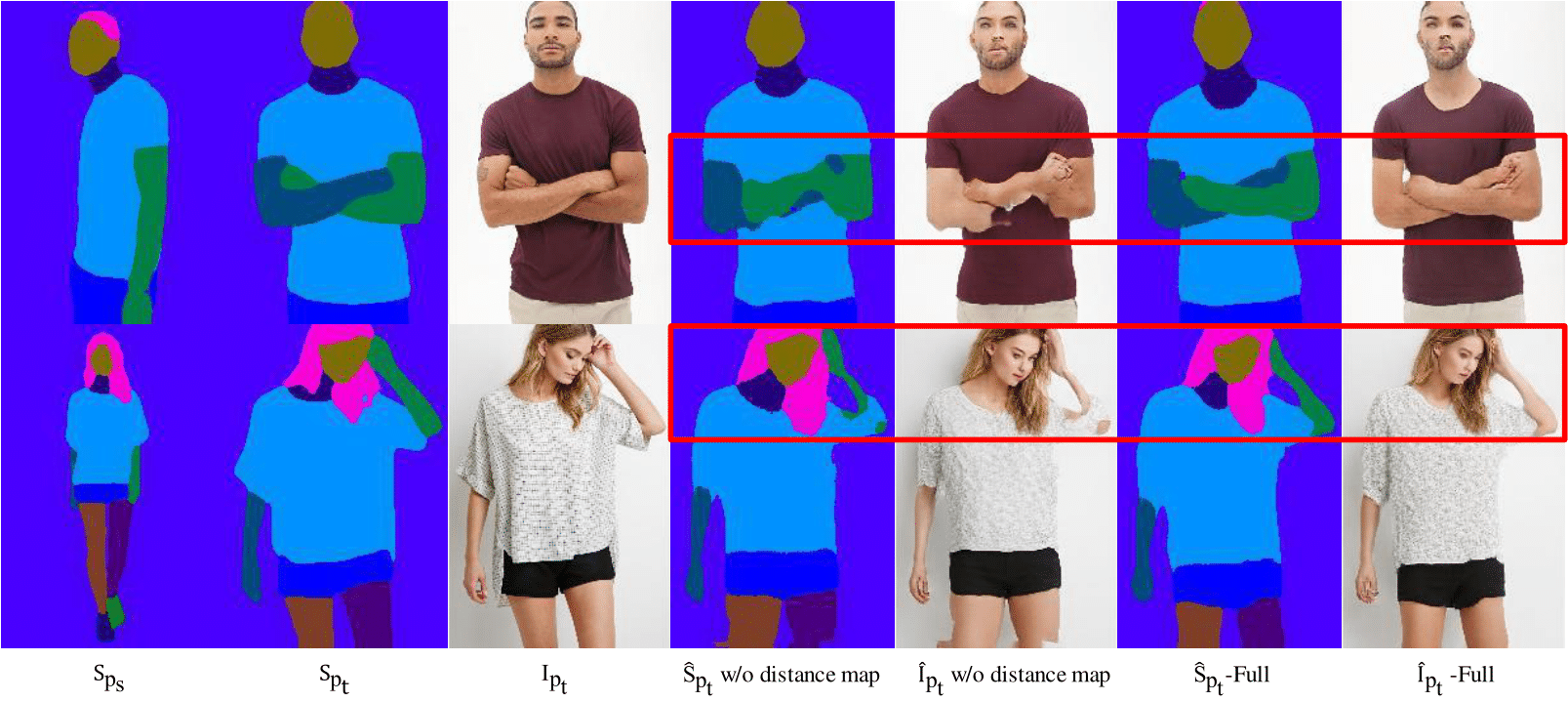}
		\caption{The gain of the distance map to the image generation.}
		\label{fig:dpcpr}
		\vspace{-13.5pt}
	\end{figure*}

	It may be not enough to use only keypoints skeleton as pose representations to generate semantic maps, especially when the poses are complex or rare in the dataset. The introduction of distance maps makes pose representation more robust to various poses.
	We generate 12 lines \{$L_m|_{m=1}^{12}$\} distance map with 18-channels keypoint heat maps to represent the body skeleton. Each skeleton generates one channel distance map, thus we can generate a 12-channels distance map $\{M_m|_{m=1}^{12}\}$, which has the same width and height of the source image. The values in $(x, y)$ of each $M_m$ is calculated by the smallest distance between the point $(x, y)$ and the skeleton $L_m$. The m-\textit{th} distance map can be obtained by:
	\vspace{-3pt}
	\begin{equation}
	\setlength{\abovedisplayskip}{5pt}
	\setlength{\belowdisplayskip}{5pt}
	M'_m(x,y) =\underset{(x',y')\in L_m}{\min}\{\sqrt{(x-x')^2+(y-y')^2}\},
	\end{equation}
	{where $(x',y')$ denotes the point on the skeleton $L_m$.}
	{Here, we further normalize these values by introducing a negative parameter {$\kappa$}. The final distance map is then defined as:}
	\vspace{-3pt}
	\begin{equation}
	\setlength{\abovedisplayskip}{5pt}
	\setlength{\belowdisplayskip}{5pt}
	M_m(x,y) = exp (\kappa * M'_m(x,y)),
	\end{equation}
	{where $m \in \{1,2,3...12\}$ represents the m-\textit{th} skeleton.}
	{In this way, the closer the point to the skeleton, the larger the value is. Thus it can well model the body structure. The process of distance map generation is shown in Fig. \ref{fig:distance map}}. The experimental results on DeepFashion show that without using the distance map, the mIOU of the predicted semantic maps is 0.520, while when using the distance map, the mIOU is 0.539. By using the distance map, our method can generate plausible semantic parsing maps even though the poses are complex or rare (\eg, crossed hand in Figure~\ref{fig:dpcpr}), which will further benefit the later target image generation.
	\vspace{-2pt}
	\subsection*{D. More Qualitative Results}
\begin{figure*}[!t]
		\vspace{-13pt}
		\centering
		\includegraphics[width= \linewidth]{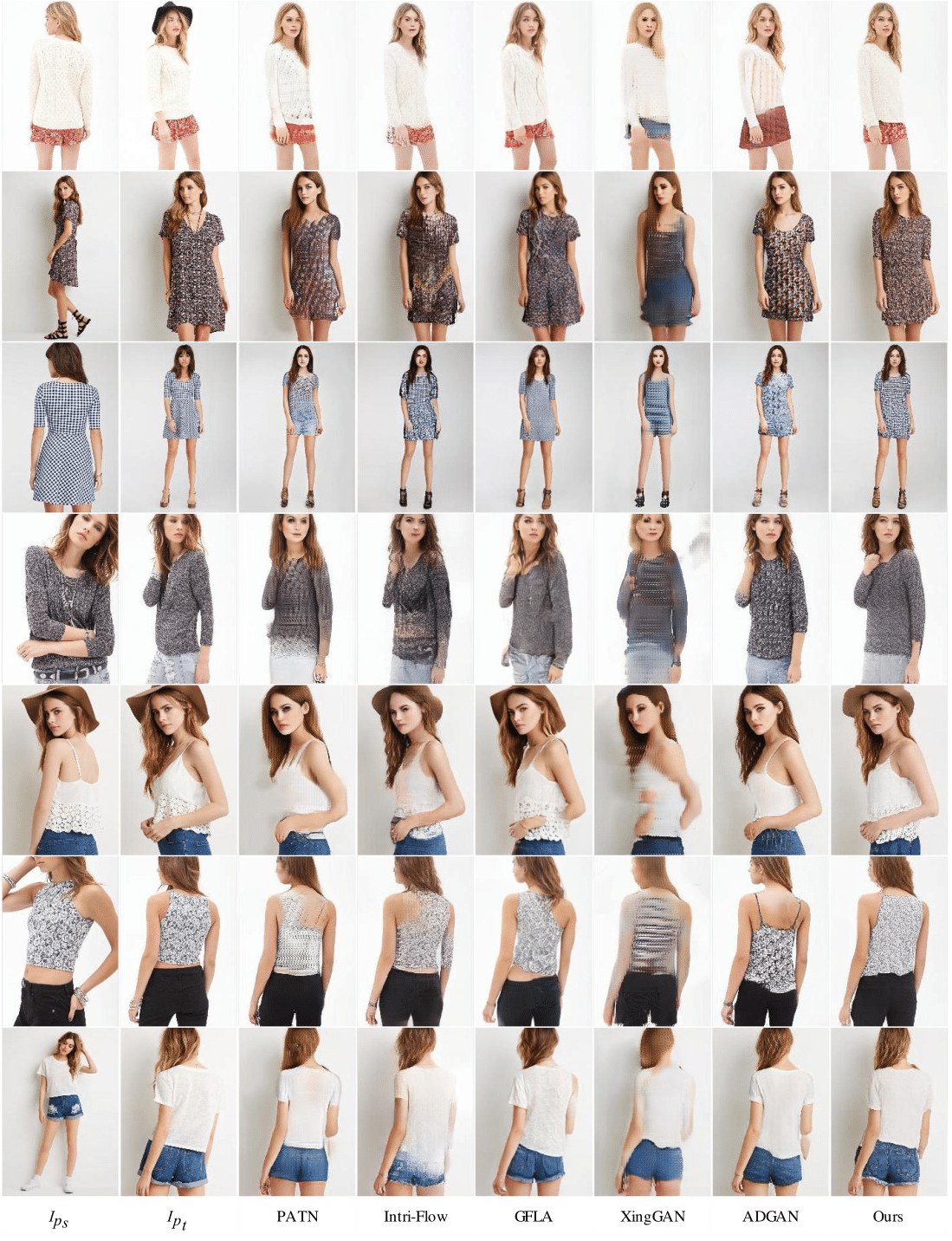}
		\caption{More visual comparison with the competing methods on DeepFashion.}
		\label{fig:comparisondp}
		\vspace{-12pt}
	\end{figure*}

\begin{figure*}[!t]
		\vspace{30pt}
		\centering
		\includegraphics[width= \linewidth]{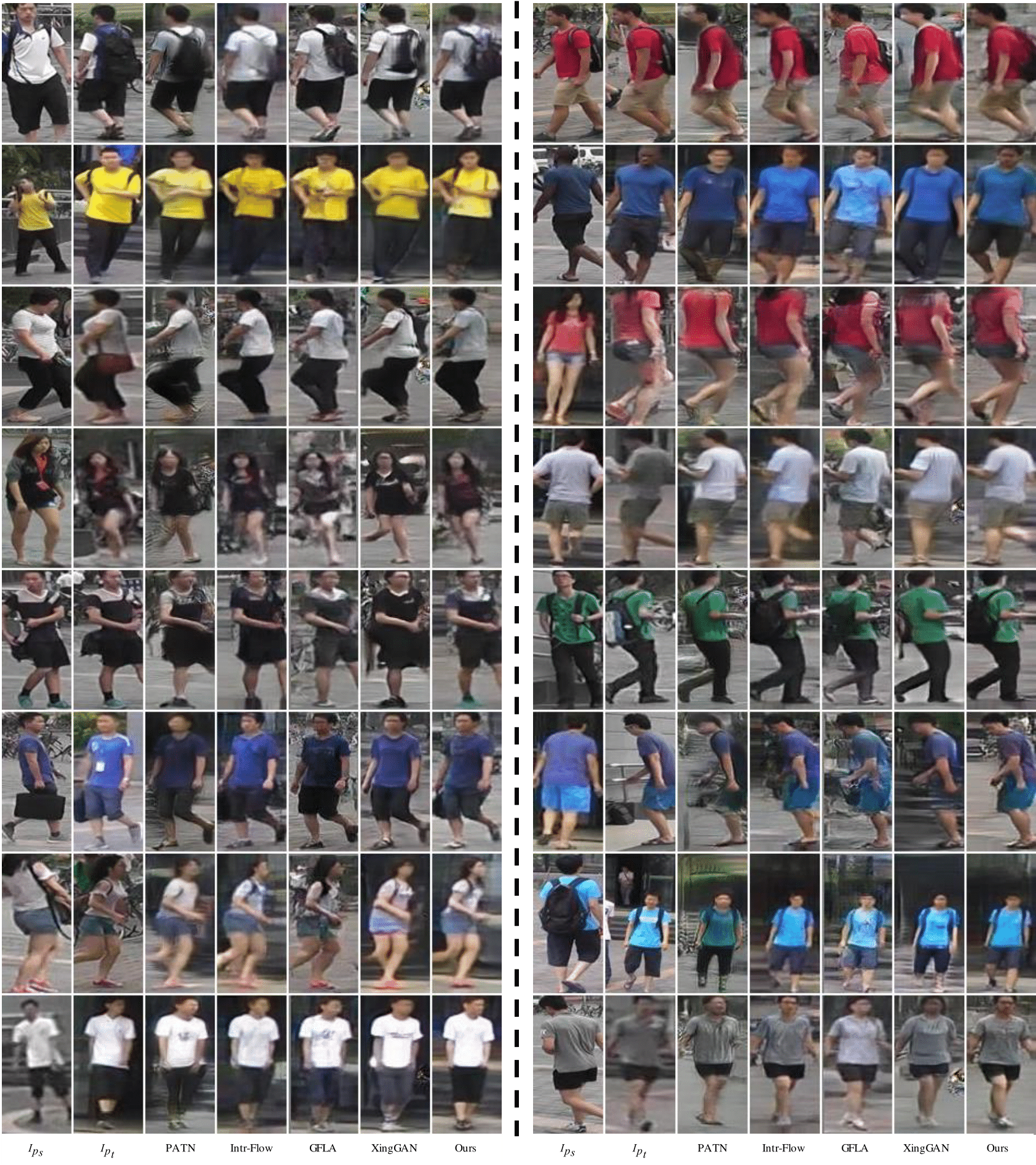}
		\caption{More visual comparison with the competing methods on Market-1501.}
		\label{fig:comparisonmk}
		\vspace{20pt}
	\end{figure*}

	In this section, we show more visual comparison with the competing methods (\ie PATN~\cite{zhu2019progressive}, Intr-Flow~\cite{li2019dense}, GFLA~\cite{ren2020deep}, XingGAN~\cite{tang2020xinggan}, ADGAN~\cite{men2020controllable}) on DeepFashion and Market-1501 in Figs.~\ref{fig:comparisondp} and \ref{fig:comparisonmk}, respectively. It can be seen that our method can generate more semantic, consistent, and photo-realistic results. Besides, we also show more visual comparison of different SPGNet variants in Fig. \ref{fig:ablation}.
	
	\begin{figure*}[!t]
		\vspace{-18pt}
		\centering
		\includegraphics[width= \linewidth]{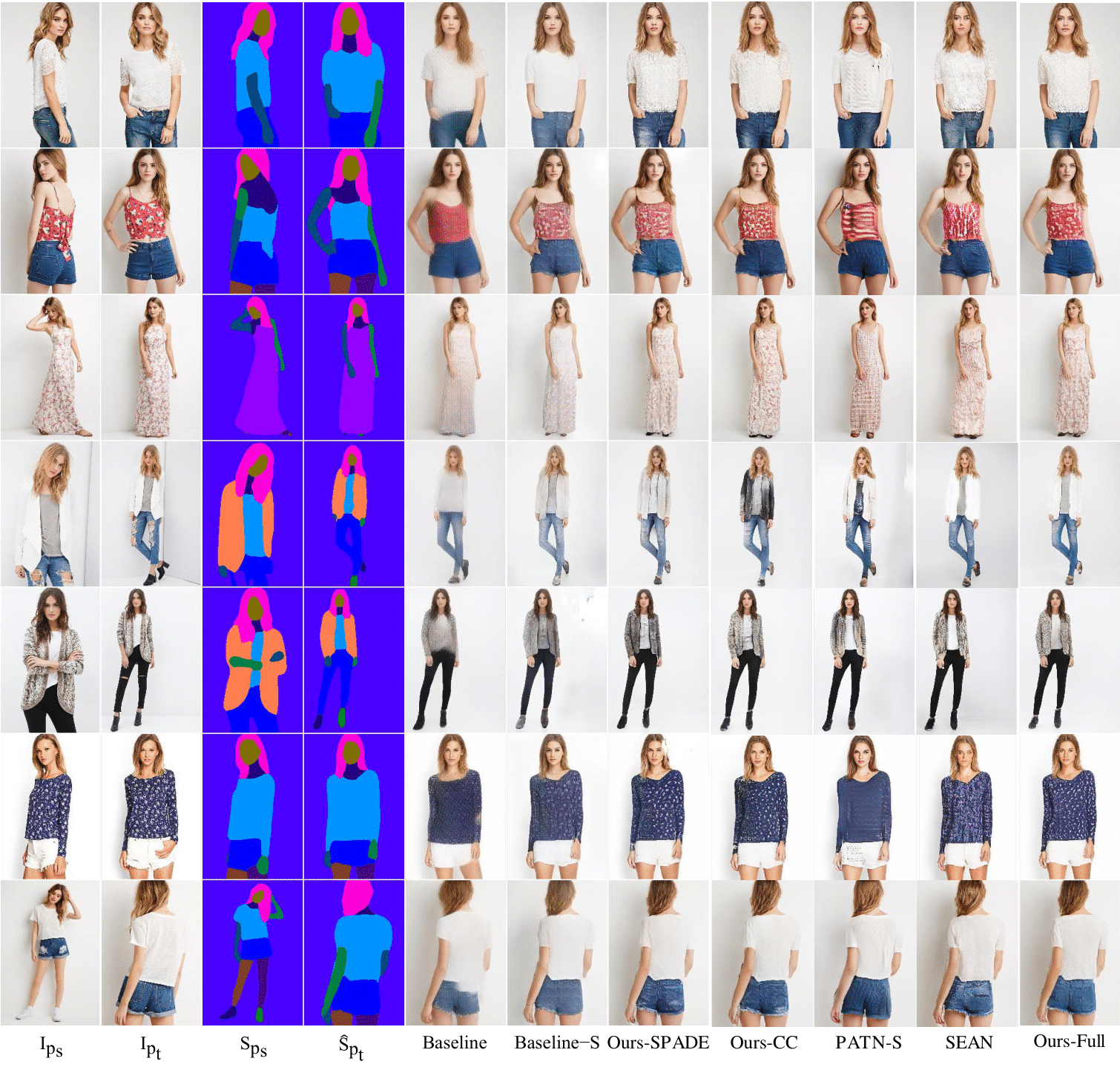}
		\caption{More visual comparison of different SPGNet variants.}
		\label{fig:ablation}
		\vspace{-15pt}
	\end{figure*}

\end{document}